\documentclass{article}

\usepackage[preprint]{neurips_2026}

\usepackage[utf8]{inputenc} 
\usepackage[T1]{fontenc}    
\usepackage{hyperref}       
\usepackage{url}            
\usepackage{booktabs}       
\usepackage{amsfonts}       
\usepackage{nicefrac}       
\usepackage{microtype}      
\usepackage{xcolor}         
\usepackage{listings}

\usepackage{graphicx}
\usepackage{amsmath}
\usepackage{amssymb}
\usepackage{array}
\usepackage{multirow}
\usepackage{tabularx}

\newcolumntype{L}[1]{>{\raggedright\arraybackslash}m{#1}}
\newcolumntype{Y}{>{\raggedright\arraybackslash}X}
\lstdefinestyle{prompttemplate}{
  basicstyle=\ttfamily\scriptsize,
  breaklines=true,
  breakatwhitespace=false,
  breakindent=0pt,
  breakautoindent=false,
  keepspaces=true,
  showstringspaces=false,
  frame=single,
  framerule=0.3pt,
  literate={—}{{\textemdash}}1
}

\title{FutureWorld: A Live Reinforcement Learning Environment for Predictive Agents\\with Real-World Outcome Rewards}

\author{%
\textbf{Zhixin Han}$^{1,6\ast}$, \textbf{Yanzhi Zhang}$^{2,6\ast}$, \textbf{Chuyang Wei}$^{3,6}$, \textbf{Maohang Gao}$^{3,6}$,\\
\textbf{Xiawei Yue}$^{1,6}$, \textbf{Kefei Chen}$^{5,6}$, \textbf{Yu Zhuang}$^{4,6}$, \textbf{Haoxiang Guan}$^{3,6}$, \textbf{Jiyan He}$^{6}$,\\
\textbf{Jian Li}$^{5}$, \textbf{Yitong Duan}$^{6\dagger}$, \textbf{Yu Shi}$^{6\dagger}$, \textbf{Mengting Hu}$^{1\dagger}$,
\textbf{Shuxin Zheng}$^{6\dagger}$\\[0.6ex]
$^{1}$\textbf{College of Software, Nankai University}\\
$^{2}$\textbf{Academy of Mathematics and Systems Science, Chinese Academy of Sciences}\\
\scalebox{0.97}[1]{$^{3}$\textbf{School of Computer Science and Technology, University of Science and Technology of China}}\\
$^{4}$\textbf{Institute of Automation, Chinese Academy of Sciences}\\
$^{5}$\textbf{IIIS, Tsinghua University} \quad
$^{6}$\textbf{Zhongguancun Academy, Beijing, China}\\[0.5ex]
{\small\texttt{zhixinhan@mail.nankai.edu.cn, zhangyanzhi20@mails.ucas.ac.cn,}}\\
{\small\texttt{duanyitong@zgci.ac.cn, shiyu@bza.edu.cn, mthu@nankai.edu.cn, sz@zgci.ai}}
}

\begin{document}

\maketitle

\begingroup
\renewcommand{\thefootnote}{\fnsymbol{footnote}}
\footnotetext[1]{These authors contributed equally to this work.}
\footnotetext[2]{Corresponding authors.}
\endgroup

\begin{abstract}
Live future prediction refers to the task of making predictions about real-world events before they unfold. This task is increasingly studied using large language model-based agent systems, and it is important for building agents that can continually learn from the real world. It can provide a large number of prediction questions grounded in diverse real-world events, while preventing answer leakage. To leverage the advantages of future prediction, we present FutureWorld, a live agentic reinforcement learning environment that closes the training loop between prediction, outcome realization, and parameter updates. Specifically, we modify and extend \texttt{verl-tool}, resulting in a new framework that we call \texttt{verl-tool-future}. Unlike standard reinforcement learning training frameworks that rely on immediate rewards, \texttt{verl-tool-future} stores prediction-time rollouts, backfills rewards after real-world outcomes become available, and then replays the completed trajectories for policy update. Across three open-source agents, successive FutureWorld training rounds lead to consistent improvements in prediction accuracy, probabilistic scoring, and calibration, demonstrating that delayed real-world outcome feedback can serve as an effective reinforcement learning signal.
\end{abstract}

\section{Introduction}

Live future prediction refers to the task of making predictions about real-world events before they unfold, where the results are unknown at prediction time but can be verified based on subsequent real-world outcomes \cite{FutureX,FutureX-Pro,ForecastBench}. Recent works \cite{FutureX,FutureX-Pro,ForecastBench,Turtel,Metaculus,Echo,Mantic,MiroFlow} increasingly study this task using large language model (LLM)-based agent systems, often in the form of agentic prediction pipelines that retrieve information, reason, and generate predictions. This task naturally supports the development of agent systems that make predictions, observe realized outcomes, and update their policy accordingly. In this way, they can remain aligned with evolving real-world dynamics instead of relying solely on static knowledge acquired during training. It is important for building agents that can continually learn from real-world feedback, adapt to changing conditions, and remain reliable in the dynamic real world.

Recent works on LLM agents have explored interactive environments for agentic capabilities, including web navigation, application use, and knowledge-work automation \cite{WebArena,AppWorld,WorkArena,CLIN,LOOP}. Similarly, live future prediction involves a temporal interaction loop in which agents make predictions about unresolved real-world events, receive outcome-grounded feedback after those events are realized, and improve their policies over time. This naturally motivates treating it as a learning environment. Such an environment should satisfy three requirements. \textbf{First}, the stream of questions should remain live, because new prediction targets continually emerge from ongoing real-world developments, and a static collection of historical questions cannot meet the need for live continual learning. \textbf{Second}, the learning signal should be grounded in realized outcomes, rather than relying solely on proxy signals such as process rewards, so that optimization remains aligned with the true objective of prediction. \textbf{Third}, training should include the agent's own information retrieval and analysis, because in realistic future prediction the agent must learn not only how to make predictions from given information, but also what information to seek and how to interpret it. In this way, live future prediction defines a learning environment in which agents can continually adapt their policies through real-world feedback and continually update their understanding of an evolving world.

Prior works have explored future prediction from different aspects. Existing works \cite{FutureX,ForecastBench,FutureX-Pro} take \textbf{an important step} by providing live evaluation infrastructures that continually introduce new prediction questions over time. \textbf{Another step} is to place learning and agentic rollouts inside a live setting. \citet{Echo} moves in this direction by combining reinforcement learning (RL) with agentic rollouts in a live prediction setup. However, its learning signal is based on rubric-based process rewards rather than realized outcomes, leaving a gap between the training signal and the true objective of prediction. \textbf{A further step} is to optimize learning directly against realized outcomes. Recent works \cite{Turtel,Mantic} move toward this objective by exploring outcome-based RL. However, they do so only on static datasets of previously resolved questions, where the evidence available for prediction is fixed in advance. As a result, the agent is not truly trained to search for, select, and interpret evidence in a live environment. Taken together, prior works have explored different important aspects of live future prediction, but no existing work has yet established it as a unified learning environment.

A closer look at live future prediction reveals several properties that make it particularly well suited as a learning environment. Feedback can be derived directly from real-world outcomes, without requiring human annotation. This greatly reduces cost and makes it possible to scale training over a large number of prediction questions. In addition, the diversity of the real world allows prediction questions to span many domains. Notably, data leakage is prevented by construction.

\begin{figure}
  \centering
  \includegraphics[width=1\columnwidth]{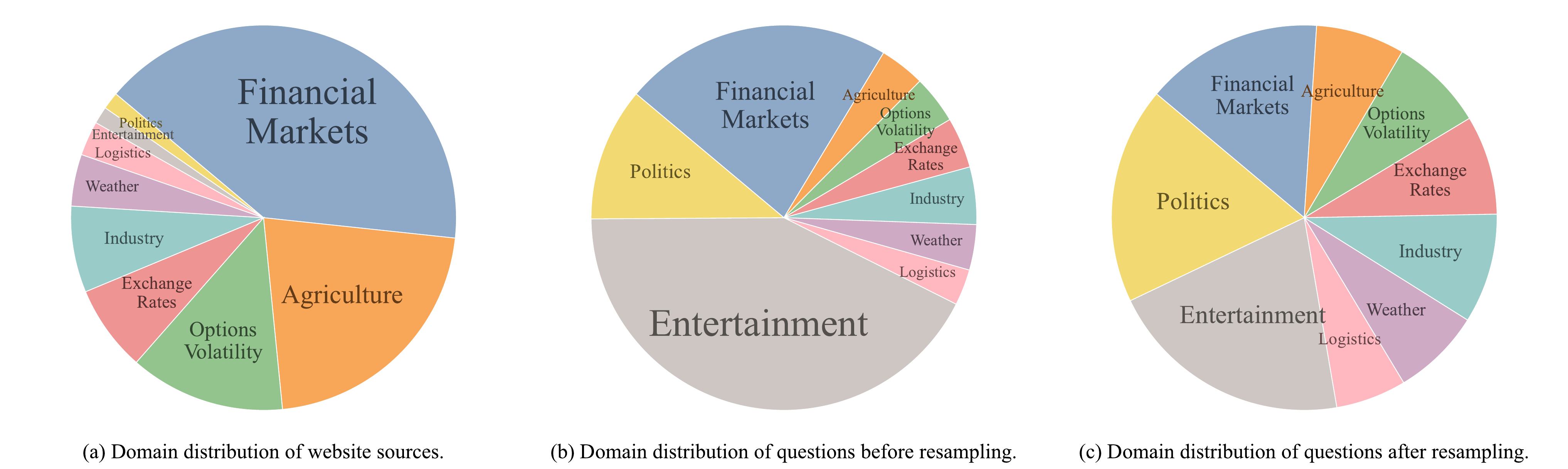}
  \caption{Domain distributions of website sources (a), questions before resampling (b), and questions after resampling (c). After resampling, questions are more evenly distributed across domains.}
  \label{distribution}
\end{figure}

To leverage the advantages of live future prediction, we present FutureWorld, a live agentic RL environment that closes the training loop between prediction, outcome realization, and parameter updates. Each day, the system automatically generates a large number (2,047.29 on average, see Section~\ref{resample}) of prediction questions from a broad set of carefully selected, high-value event sources spanning many domains, as illustrated in Figure~\ref{distribution} (a). The generated questions are first filtered to remove low-quality items, after which the remaining questions are resampled to balance domain proportions while reducing the number of similar questions within each domain. For each prediction question, the agent executes a rollout that may involve issuing search queries, reading retrieved information, reasoning, and producing a prediction, while the environment records the trajectory. The ground-truth outcome is provided later, when the event resolves, at which point it is matched to the corresponding stored trajectory and used to compute a reward. To support this delayed-feedback loop, we modify and extend \texttt{verl-tool} \cite{Verl-Tool}, and introduce the resulting framework as \texttt{verl-tool-future}. This proposed framework decouples prediction-time rollout collection from later outcome retrieval, reward backfilling, and policy update. The accumulated rewards are then used to update the model's parameters. The entire pipeline runs autonomously on a daily cycle.

In our environment, we take three open-source base language models and train them using outcome-based RL. The results show that training is effective. In summary, our contributions are as follows:

\begin{itemize}
\item We introduce FutureWorld, an agentic RL learning environment for live future prediction. To the best of our knowledge, it is the first open learning environment in which agents learn directly from the outcomes of their own predictions about the future.
\item We propose \texttt{verl-tool-future}, our modified and extended version of \texttt{verl-tool} for delayed-feedback RL in live future prediction.
\item We provide empirical evidence that outcome-based RL in a live environment can improve prediction ability over time. Open-source models trained in FutureWorld achieve progressively better performance across successive training days.
\end{itemize}

\section{Related work}

\begin{figure}
  \centering
  \includegraphics[width=1\columnwidth]{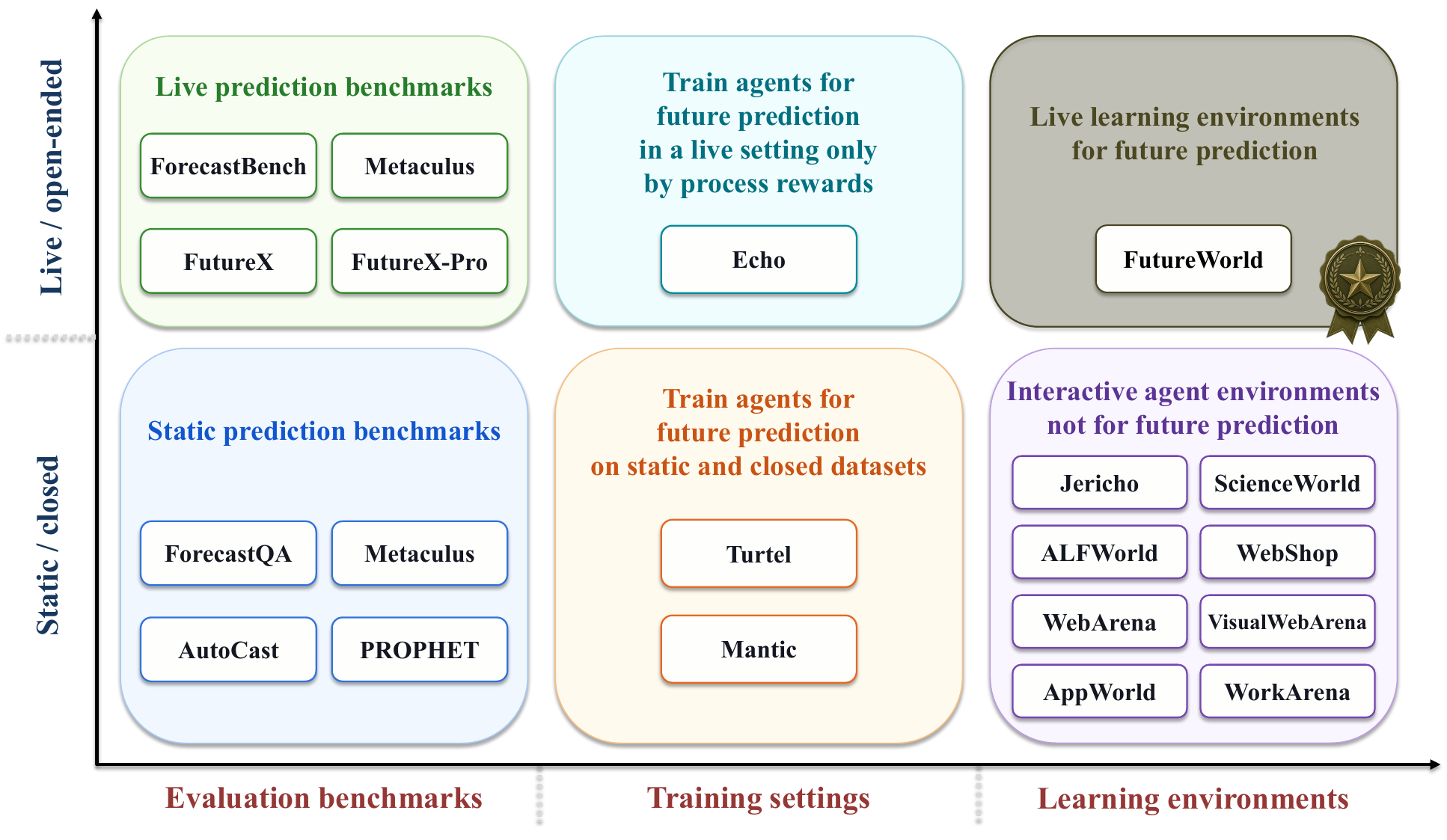}
  \caption{FutureWorld fills the gap for predictive agents.}
  \label{teaser}
\end{figure}

Prior works fall into three lines. Future prediction benchmarks evaluate models on unresolved real-world events. Recent methods for training predictive agents have improved performance, but often rely on static non-live datasets, pre-retrieved information, or indirect process-based reward signals. Agent environments, meanwhile, highlight the importance of interaction with timely and reproducible feedback, but are usually not built for live future prediction with direct outcome-grounded learning signals. Figure~\ref{teaser} highlights the remaining gap.

\subsection{Benchmarks for future prediction}

Future prediction benchmarks evaluate models on real-world events. Earlier benchmarks such as ForecastQA \cite{ForecastQA} and AutoCast \cite{AutoCast} use historical prediction questions, while PROPHET \cite{PROPHET} further emphasizes inferability by constructing prediction questions paired with supporting news evidence. More recent efforts have moved prediction evaluation closer to live settings. ForecastBench \cite{ForecastBench} and FutureX \cite{FutureX} introduce automatically updated benchmarks of unresolved prediction questions, while FutureX-Pro \cite{FutureX-Pro} extends this live paradigm to higher-value vertical domains.

\subsection{Training agents for future prediction}

Existing methods have already explored the problem of improving future prediction ability in LLM-based agent systems. \textbf{Some works} apply outcome-based RL to static historical datasets of already resolved prediction questions. \citet{Turtel} demonstrate that fine tuning with negative Brier score as a reward signal yields substantial accuracy gains on a 14B parameter model trained over 110,000 resolved Polymarket events,\footnote{\url{https://polymarket.com}} but the agent operates only on a fixed prompt that contains pre-collected information, rather than acquiring information through its own web search process, so the information gathering policy remains entirely untrained. \citet{Mantic} apply Group Relative Policy Optimization (GRPO) \cite{GRPO} to a 120B-parameter model using the Brier score reward on approximately 10,000 historical binary prediction questions with resolved outcomes, achieving notable improvements on the Metaculus benchmark \cite{Metaculus}. However, its research phase is also performed before training. \textbf{Other work}, exemplified by \cite{Echo}, operates in a live setting with agentic rollouts and a daily rolling cycle, but uses rubric-based process rewards rather than direct outcome-based rewards, introducing an indirection between the training signal and the target objective. Across these efforts, no existing system simultaneously trains on outcome-derived rewards, performs agentic information retrieval inside the training loop, and operates over a live, rolling stream of questions.

\subsection{Agent environments}

Agent environments have established the value of interaction and feedback for training and evaluating large language agents. Prior works have studied a range of environments for interactive agent tasks with timely feedback and reproducible evaluation. Early text-based settings such as Jericho \cite{Jericho} provide interactive fiction games for language-conditioned action. ScienceWorld \cite{ScienceWorld} focuses on grounded scientific reasoning in an interactive text environment, and ALFWorld \cite{ALFWorld} connects text-based interaction with simulated embodied household tasks. Other works study web and software interaction in more realistic yet still static and closed settings. WebShop \cite{WebShop} formulates grounded web interaction as an online shopping task, WebArena \cite{WebArena} recreates functional websites in a reproducible sandbox, VisualWebArena \cite{VisualWebArena} extends this paradigm to visually grounded web tasks, AppWorld \cite{AppWorld} builds a controllable ecosystem of apps populated with fictitious users, and WorkArena \cite{WorkArena} targets browser-based knowledge-work tasks in enterprise software. However, all of these environments remain static (non-live) and closed, which limits their ability to expose agents to open-ended real-world dynamics.

\section{FutureWorld environment}

\subsection{Problem definition}
\label{definition}

We formulate live future prediction as a delayed-feedback interaction problem. Each instance is a prediction prompt $q$ about a future event whose outcome is unknown at prediction time. At prediction time $t_q^{\mathrm{pred}}$, an agent interacts with external information sources through a sequence of search actions and observations, and then produces a final probability estimate for whether the target event will occur. In this formulation, each live future prediction instance is converted into a binary classification problem. For the $k$-th stochastic rollout of question $q$, FutureWorld represents the completed trajectory as
\begin{equation}
\begin{aligned}
\tau_{q,k} =
\bigl(&q,\,
t_q^{\mathrm{pred}},\,
a_{q,k,1}, o_{q,k,1}, \ldots,
a_{q,k,m}, o_{q,k,m},
\hat{\pi}_{q,k},\,
t_q^{\mathrm{resolve}},\,
z_q,\,
r_{q,k}
\bigr),
\end{aligned}
\end{equation}
where $a_{q,k,i}$ is the $i$-th search action, $o_{q,k,i}$ is the corresponding search observation, $m$ is the number of search steps in the rollout, $\hat{\pi}_{q,k}\in\left[0,1\right]$ is the final probability estimate, $t_q^{\mathrm{resolve}}$ denotes the scheduled time at which FutureWorld attempts to retrieve the outcome of $q$, $z_q\in\{0,1\}$ is an indicator of whether the target event actually occurs, and $r_{q,k}$ is the delayed trajectory-level reward.

The trajectory is completed in two distinct stages. Before the event resolves, FutureWorld stores the prediction-time prefix containing $q$, $t_q^{\mathrm{pred}}$, the action-observation sequence, and $\hat{\pi}_{q,k}$. At the scheduled resolution time $t_q^{\mathrm{resolve}}$, FutureWorld attempts to retrieve the event outcome. If the outcome is available, the environment backfills the realized label
\begin{equation}
z_q =
\begin{cases}
1, & \text{if the target event occurs},\\
0, & \text{otherwise}.
\end{cases}
\end{equation}
It then computes the delayed reward $r_{q,k}=R(\hat{\pi}_{q,k}, z_q)$. This schema makes FutureWorld a live environment with timestamps, tool interaction, delayed outcome labels, and reward backfilling.

\subsection{Environment overview}

\begin{figure}
  \centering
  \includegraphics[width=1\columnwidth]{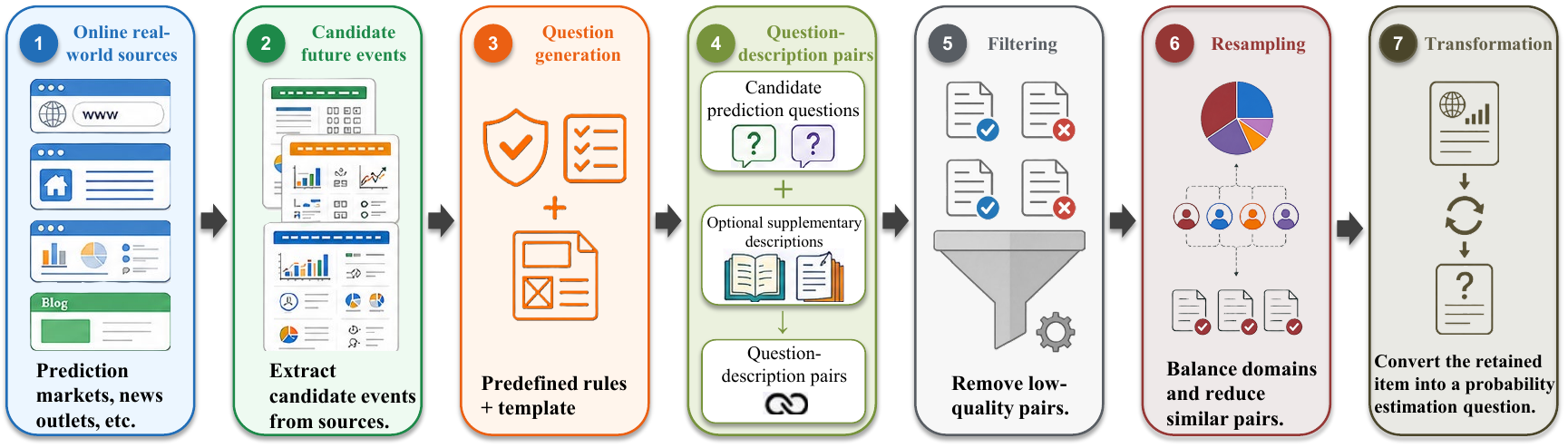}
  \caption{Overview of the FutureWorld pipeline for constructing prediction prompts.}
  \label{prompt}
\end{figure}

Building on the definition above, FutureWorld implements a live environment that supports the complete pipeline from question generation to delayed reward assignment. As illustrated in Figure~\ref{prompt}, the process begins by collecting data from a broad set of public online sources about candidate future events, and then constructs questions using predefined rules or templates and, when applicable, associates them with supplementary descriptions that provide additional background information. These pairs are further processed through filtering and resampling so that the final retained set is higher quality, more balanced across domains, and has lower within-domain question similarity. The retained questions are then converted into final prediction prompts, where the agent is asked to estimate the probability that the specified future event will actually occur.

\begin{figure}
  \centering
  \includegraphics[width=1\columnwidth]{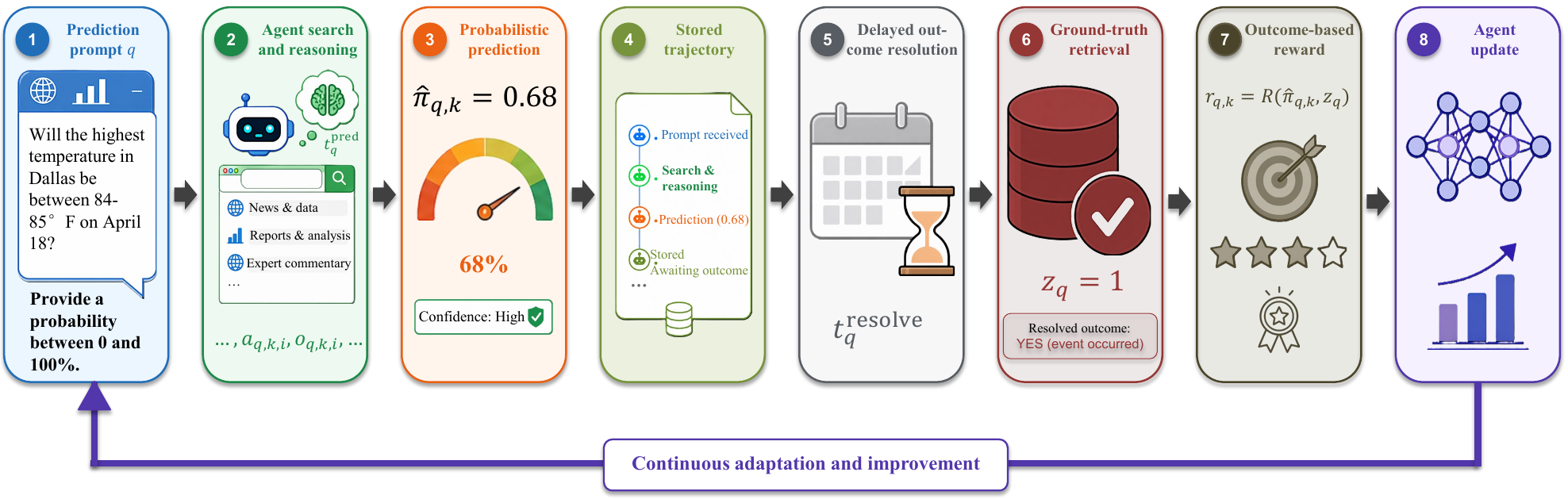}
  \caption{Overview of the FutureWorld training loop.}
  \label{train}
\end{figure}

Once deployed in the environment, as shown in Figure~\ref{train}, an agent receives a prediction prompt $q$ at prediction time $t_q^{\mathrm{pred}}$, interacts with external information sources through a sequence of search actions and observations $\{(a_{q,k,i}, o_{q,k,i})\}_{i=1}^{m}$, and then produces a final probability estimate $\hat{\pi}_{q,k}$. FutureWorld records this prediction-time process as the prefix of a rollout trajectory $\tau_{q,k}$. At the scheduled resolution time $t_q^{\mathrm{resolve}}$, when the corresponding real-world outcome is expected to be available, the environment attempts to retrieve the ground truth, backfills the resolved label $z_q$, and computes the outcome-based reward $r_{q,k}$. In this way, FutureWorld serves as a continuously refreshed delayed-feedback environment that maintains a persistent stream of future prediction questions and closes the loop among agent interaction, realized outcomes, and policy improvement.

\subsection{Data sources}

FutureWorld begins by collecting candidate future events from a broad set of online sources through automated network requests.\footnote{FutureWorld only uses publicly accessible sources. The collected data are used only for academic research, and not for any illegal or commercial purpose.} We maintain a pool of 72 websites. These source websites cover a wide range of domains. Figure~\ref{distribution} (a) shows the domain distribution of the source websites. Most of these websites cover consequential real-world developments.

\subsection{Construction of question-description pairs}

After collecting data from source websites, FutureWorld instantiates candidate prediction questions using predefined rules or templates. These questions are formulated as binary prediction questions that ask whether a specific event will occur. For example, a question may take the form \textit{Will the highest temperature in Dallas be between 84-85°F on April 18?} Each question may also be paired with an optional supplementary description to provide additional background information. For some websites, the description can be directly extracted from the source page, whereas for websites that periodically publish a fixed set of data types, we pre-generate type-level descriptions with the assistance of an LLM, GPT-5.4.\footnote{\url{https://openai.com/index/introducing-gpt-5-4}}

\subsection{Question-description pair filtering}

Based on our exploration, we identify three criteria that high-quality prediction questions should satisfy. \textbf{First}, it should be objectively resolvable. The outcome should be verifiable from publicly accessible evidence. \textbf{Second}, it should concern a meaningful real-world outcome. \textbf{Third}, it should be safe for public release. The question should exclude sensitive, harmful, or otherwise inappropriate content. To enforce all these quality requirements, FutureWorld applies a filtering stage.

To operationalize the three criteria above, we design three filters for quality control. Each filter asks the LLM to judge whether a question-description pair violates the corresponding eligibility criteria. The optional description included in the input helps the LLM assess eligibility more reliably even without web search, thereby reducing the cost. A pair is removed if any filter flags it as ineligible. We use \texttt{bytedance-seed/seed-1.6} as the filtering model.\footnote{\url{https://seed.bytedance.com/en/seed1\_6}}

\subsection{Question-description pair resampling}
\label{resample}

After filtering, FutureWorld applies resampling to the remaining question-description pairs. It is designed to balance the proportions across different domains while reducing question similarity within each domain. The target number of questions retained after resampling can be manually specified. The detailed resampling procedure is provided in Appendix~\ref{details}.

After performing the resampling procedure, we obtain the final set of questions. In practice, before resampling, we typically obtain around 2,047.29 questions on average across seven observations. We set the target number of questions retained to 500. Figure~\ref{distribution} (b) and (c) show the domain distributions before and after resampling, respectively. As shown in the figure, the distribution of questions across domains becomes more balanced after resampling.

\subsection{Prompt construction for probabilistic prediction}

The resampled set is then passed to the prompt construction stage. FutureWorld applies a predefined prompt template to convert each retained binary question into a probability estimation question. This probabilistic formulation allows the agent to express uncertainty explicitly and provides a richer, more fine-grained supervision signal for learning. Although some questions are paired with supplementary descriptions in earlier stages to support filtering and resampling, these descriptions are omitted from the final prompts given to the agent, so that the agent is not given additional hints.

\subsection{Ground-truth retrieval}

Outcome retrieval is implemented as a set of source-specific resolvers rather than a single universal rule, since different sources expose outcomes through different interfaces and encode resolution states in different formats. For sources supported by \texttt{AkShare}~\cite{AkShare}, the corresponding resolver queries \texttt{AkShare} to obtain the structured data used for outcome resolution. Each question record stores the information needed to route the question to the appropriate resolver and verify the returned outcome, including the source website, source URL, expected resolution time, source-specific metadata when available, and other fields. If the source has not yet published a valid outcome, or if the returned information cannot be reliably matched to the original question, the question is marked as unresolved and excluded from scoring.

\subsection{Training loop through verl-tool-future}

To decouple data collection, prediction-time rollout generation, outcome acquisition, and model updating, we modify \texttt{verl-tool} \cite{Verl-Tool} to support storing LLM rollouts. We refer to our modified framework as \texttt{verl-tool-future}.

At 20:00 on day $t$, we use the FutureWorld environment to obtain a batch of prediction questions whose outcomes are expected to be resolved on the following day, i.e., day $t+1$. The LLM agent is required to perform at least one web-search action before giving its final prediction, and the prediction-time trajectory prefix defined in Section~\ref{definition} is saved. At 20:30 on day $t+1$, FutureWorld attempts to retrieve answers for the batch of prediction questions issued on day $t$. For questions whose outcomes are successfully obtained, it backfills the resolved label $z_q$ and the corresponding reward $r_{q,k}$ into the stored trajectories. The completed trajectories are then used to update the model parameters through reinforcement learning.

FutureWorld runs every day to collect new prediction questions and attempt to retrieve outcomes for the questions issued yesterday. However, real-world events and public reporting are not perfectly synchronized. Even when an event has already resolved, its outcome may not yet be available on the corresponding source website because of delays. Furthermore, some prediction targets may be postponed or canceled. Overall, when ground-truth retrieval is performed in the evening, some previous prediction questions may still remain unresolved. Across five consecutive days of observation, an average of 35.65\% of questions do not have retrievable ground truth at the scheduled retrieval time. When ground truth cannot be successfully retrieved, our strategy is to discard the corresponding questions and exclude them from training.

\section{Training agents in FutureWorld}



\subsection{Reward and replay}

We define trajectory-level reward as the negative Brier loss \cite{Mantic,Turtel}:
\begin{equation}
      r_{q,k} = -\left(\hat{\pi}_{q,k} - z_q\right)^2 .
\end{equation}
If the agent fails to output a valid probability in $\left[0,1\right]$, we set $r_{q,k}=-1$, the lowest possible reward.

We optimize the policy with GRPO~\cite{GRPO}. During rollout generation, the agent's search queries are treated as policy actions, while the information returned by the search tool is treated as external observations. These tool observations are masked in the policy loss because they are produced by the external environment rather than by the policy itself.

\subsection{Experiments}

\subsubsection{Performance gains across consecutive training days}

\begin{figure}
  \centering
  \includegraphics[width=1\columnwidth]{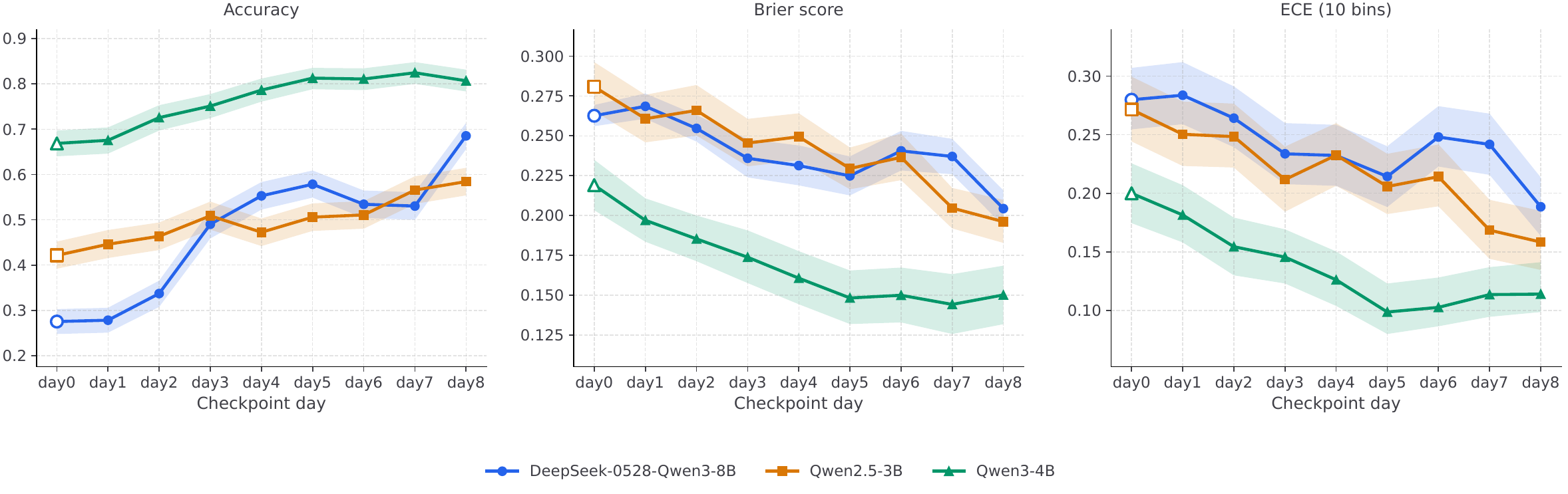}
  \caption{Prediction performance across model checkpoints saved on different days. Shaded regions indicate 95\% bootstrap confidence intervals.}
  \label{results}
\end{figure}

We save the resulting model checkpoint after each day of training. After training for 8 consecutive days, we evaluate all daily checkpoints on the same set of 500 prediction questions to ensure a consistent comparison. All nine checkpoints (day0-day8) are evaluated on the same day. Figure~\ref{results} summarizes the evaluation results. It reports accuracy after converting each probabilistic estimation into a binary prediction using a 0.5 threshold. When the predicted probability is 0.5, we treat it as a positive prediction, i.e., the model is considered to predict that the specified event will occur in the future. Accuracy generally improves over the consecutive training days. Figure~\ref{results} also reports the Brier score, which directly evaluates the quality of the probabilistic predictions. For the Brier score, lower values indicate better predictions. Furthermore, we use 10 equal-width probability bins to compute expected calibration error (ECE) \cite{ECE}. ECE decreases gradually over time. All results indicate that RL training in FutureWorld can improve agents' ability to predict future events.

\subsubsection{Domain-wise performance gains after FutureWorld training}

\begin{figure}
  \centering
  \includegraphics[width=1\columnwidth]{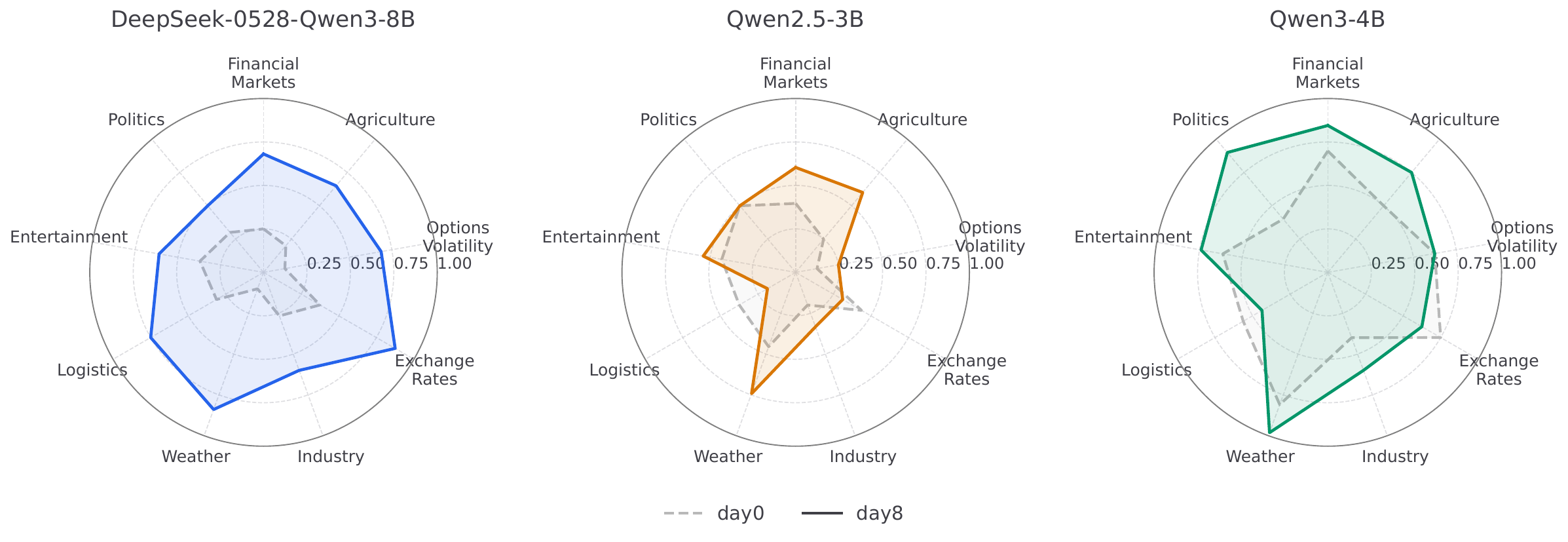}
  \caption{Domain-wise prediction performance before and after FutureWorld training.}
  \label{radar}
\end{figure}

Figure~\ref{radar} provides a domain-wise comparison between the initial models and the checkpoints after 8 days of FutureWorld training. The figure shows that the day-8 checkpoints generally outperform the initial models across most domains, indicating that the gains are not driven by a single domain of questions. This suggests that FutureWorld training improves the agents' general ability to gather information and reason, rather than merely adapting to a narrow domain-specific pattern.

\subsubsection{Generalization beyond binary questions}

\begin{figure}
  \centering
  \includegraphics[width=1\columnwidth]{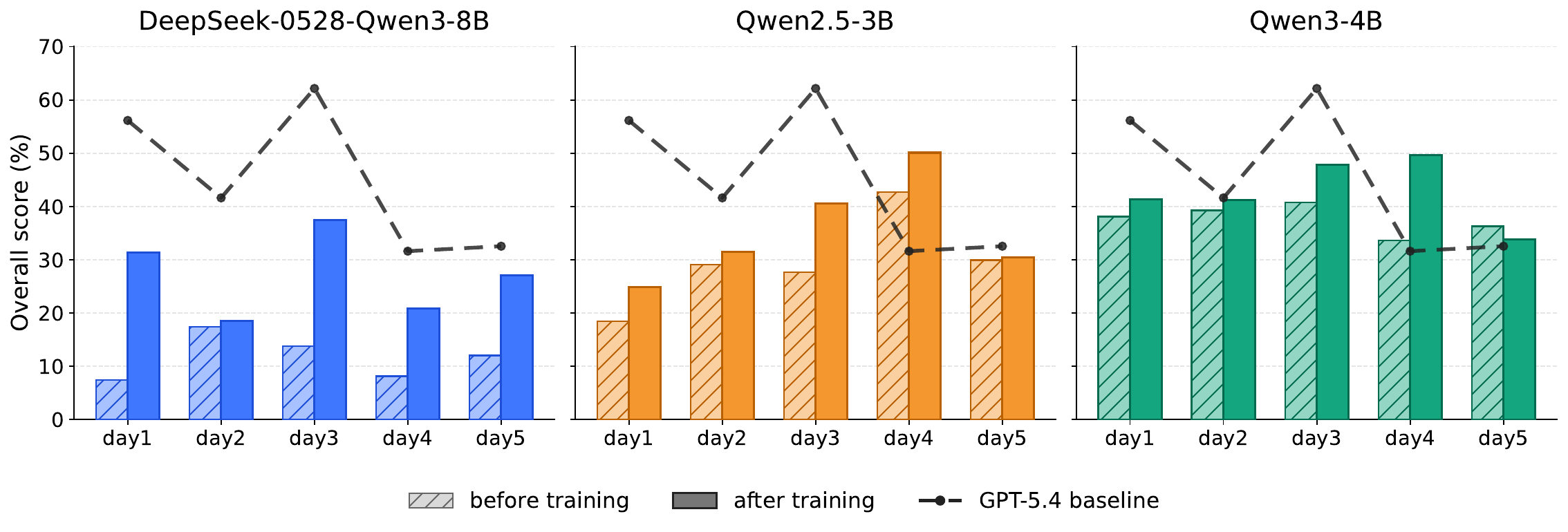}
  \caption{Day-level overall scores on the FutureWorld daily benchmark over five consecutive days. Bars compare each agent before and after FutureWorld RL training, and the dashed line shows the GPT-5.4 web-search baseline.}
  \label{overfit}
\end{figure}

Our RL training is conducted on binary questions. We therefore further examine whether the acquired predictive ability transfers beyond this binary format. To this end, we design the FutureWorld daily benchmark as a more general-purpose benchmark for evaluating predictive agents, as described in detail in Appendix~\ref{benchmark}. As summarized in Table~\ref{types}, the benchmark covers four question types: binary choice, simple multiple choice, difficult multiple choice, and numeric prediction. Representative examples of each type are shown in Table~\ref{examples}. Unlike the binary probabilistic prompts used during training, the daily benchmark requires agents to answer in heterogeneous formats. The prompt templates used for these four benchmark formats are provided in Appendix~\ref{templates}. This allows us to test whether RL improves the ability to gather information and reason about uncertain future events, rather than simply overfitting to the binary prediction questions. We evaluate our trained agents on five consecutive days, and compare their performance with that of the corresponding untrained models. The results are reported in Figure~\ref{overfit}. As shown in the figure, all agents achieve performance gains after RL training, and in some cases the trained agents even outperform a strong GPT-5.4 baseline equipped with a web-search plugin. This suggests that RL training in FutureWorld indeed enhances agents' general predictive ability.

\section{Conclusion}

We introduce FutureWorld, a live environment for training predictive agents with real-world outcome rewards. Unlike prior works that rely on static collections of resolved questions or proxy process-based rewards, FutureWorld keeps the prediction stream live, places agentic search and reasoning inside the training loop, and learns directly from realized future outcomes. We develop \texttt{verl-tool-future}, our modified and extended version of \texttt{verl-tool}, which decouples prediction-time rollout from later outcome retrieval, reward backfilling, and policy update. Our experiments show that delayed real-world feedback can serve as an effective learning signal, leading to improved prediction performance over successive days. We hope FutureWorld and \texttt{verl-tool-future} can serve as useful steps toward agent systems that continually improve by making predictions, observing outcomes, and adapting to an evolving real world.

\section{Limitations \& future work}

The availability of prediction markets and odds aggregation websites such as Polymarket, Kalshi,\footnote{\url{https://kalshi.com}} Metaculus,\footnote{\url{https://www.metaculus.com/}} and Manifold\footnote{\url{https://manifold.markets/}} can provide agents with a substantial advantage, as these sites often expose crowd-aggregated beliefs about some future events. This potential source of advantage is not specific to FutureWorld. Rather, it reflects a broader design choice shared by existing agentic prediction systems \cite{MiroFlow,Echo}, which typically do not explicitly exclude prediction markets or odds aggregation websites during information retrieval. To remain consistent with this established practice, our implementation adopts the same setting. A systematic analysis of how publicly available information from prediction markets and odds aggregators affects agents' prediction policy remains an important open problem, and we view it as a direction for future work.

Another limitation concerns longer delayed feedback. In our implementation, when ground truth cannot be successfully retrieved at the scheduled retrieval time, we discard the corresponding questions and exclude them from reward computation. This strategy is simple and practical, but it inevitably wastes computation and potentially valuable supervision signals, because some questions may be discarded after one failed retrieval attempt even though their outcomes could become available on later days. Developing more effective mechanisms for incorporating such longer delayed feedback remains an important direction for future work.

\begin{ack}
This work is supported by the Zhongguancun Academy, (Grant No.s C20250210).
\end{ack}

\bibliographystyle{plainnat}
\bibliography{refs}


\appendix

\section{Resampling procedure}
\label{details}

We categorize the filtered question-description pairs into domains. Specifically, we use keyword-matching rules to determine which domain each pair belongs to. For each domain, we then select a representative subset while avoiding pairs that are semantically too similar to one another. The selected pairs are retained as the final set.

Let $M$ denote the manually specified target number of pairs to retain after the
resampling stage. Given the domain assignment, we first allocate this total budget across non-empty domains. For each domain $d$, let $\mathcal{P}_d=\{p_1,\dots,p_{N_d}\}$ denote the set of filtered pairs assigned to this domain, and let $M_d$ denote the target number of pairs to retain from this domain. The values of $M_d$ are chosen to make the domain distribution as balanced as possible, subject to the capacity constraint $M_d\leqslant N_d$, and satisfy
\[
\sum_d M_d=M,
\]
where the summation is taken over all non-empty domains.

We then perform the selection procedure separately for each domain. For each pair $p_i\in\mathcal{P}_d$, we construct a joint textual representation $s_i$ by concatenating the question with its associated description, and embed $s_i$ into a dense vector.\footnote{We encode each $s_i$ into a dense semantic representation using a pretrained embedding model, \texttt{BAAI/bge-small-en-v1.5} \cite{BGE}.} Including the description allows the resulting embedding to capture richer semantic information about the target event. We then run $K$-means clustering over these embeddings with $K=M_d$, and sample one representative pair from each cluster. This yields $M_d$ retained pairs for domain $d$.

\section{Implementation details}


All experiments are conducted on two NVIDIA A100 GPUs, each with 80 GiB of memory. We implement the delayed-feedback training loop with \texttt{verl-tool-future}, our modified \texttt{Verl-Tool}-based framework, using FSDP \cite{FSDP} as the distributed training backend and vLLM \cite{vLLM} for rollout generation. All computation is performed in BF16 precision with FlashAttention-2 \cite{FlashAttention-2} enabled. We use the AdamW optimizer \cite{AdamW} with a learning rate of $1\times 10^{-6}$ and weight decay of $0.01$. The GRPO mini-batch size is set to 32 and the per-GPU micro-batch size is set to 2. For rollout generation, we sample four stochastic trajectories for each question with temperature $1.0$ and top-$p=1.0$, so nucleus truncation is not applied \cite{nucleus}. The agent is equipped with a lightweight Google-search interface backed by the Serper API.\footnote{\url{https://serper.dev}} We train three agents initialized from \texttt{Qwen3-4B-Instruct-2507} \cite{Qwen3-4B-Instruct-2507}, \texttt{Qwen2.5-3B-Instruct} \cite{qwen2.5,qwen2}, and \texttt{DeepSeek-R1-0528-Qwen3-8B} \cite{DeepSeek-R1-0528-Qwen3-8B}. We set the number of prediction questions to 500 per day.


\section{FutureWorld daily benchmark}
\label{benchmark}

\subsection{Benchmark design}

Beyond its role as a learning environment, FutureWorld also supports a live benchmark. Compared with the questions used for RL training, it features both greater diversity in question formats and higher overall difficulty. We limit the benchmark to at most 50 questions per day. The FutureWorld daily benchmark contains four question types, summarized in Table~\ref{types}. Table~\ref{examples} provides representative examples. The prompt templates used for these four question types are provided in Appendix~\ref{templates}. Each day, the benchmark is refreshed with a new batch of prediction questions whose answers are expected to be revealed on the following day. We retrieve the ground-truth outcomes for the batch of questions released two days earlier. This design increases the fraction of questions that can be resolved.

\begin{table}
  \centering
  \small
  \caption{Question types in the FutureWorld daily benchmark.}
  \label{types}
  \begin{tabular}{@{}p{0.24\linewidth}p{0.24\linewidth}p{0.27\linewidth}p{0.14\linewidth}@{}}
    \toprule
    \textbf{Question type} & \textbf{Prediction format} & \textbf{Correct answer structure} & \textbf{Daily cap} \\
    \midrule
    Binary choice &
    Select from 2 options &
    Exactly one option &
    $\leqslant 5$ \\
    \midrule
    Simple multiple choice &
    Select from 3--4 options &
    One or multiple options &
    $\leqslant 10$ \\
    \midrule
    Difficult multiple choice &
    Select from 5--26 options &
    One or multiple options &
    $\leqslant 15$ \\
    \midrule
    Numeric prediction &
    Predict a specific value &
    A numeric value &
    $\leqslant 20$ \\
    \bottomrule
  \end{tabular}
\end{table}

\begin{table}
  \caption{Examples of the four question types in the FutureWorld daily benchmark.}
  \label{examples}
  \centering
  \begin{tabular}{>{\raggedright\arraybackslash}m{0.22\linewidth}
                >{\raggedright\arraybackslash}m{0.68\linewidth}}
    \toprule
    \textbf{Question type} & \textbf{Example} \\
    \midrule
    Binary choice &
    Will Trump visit North Korea by April 30?\newline
    A. Yes\newline
    B. No \\
    \midrule
    Simple multiple choice &
    Shimizu S-Pulse vs. V-Varen Nagasaki\newline
    A. Shimizu S-Pulse\newline
    B. Draw (Shimizu S-Pulse vs. V-Varen Nagasaki)\newline
    C. V-Varen Nagasaki \\
    \midrule
    Difficult multiple choice &
    What will Google say during their next earnings call?\newline
    A. Banana\newline
    B. Translate / Translation\newline
    C. Autonomous / Autonomously\newline
    D. Dividend\newline
    E. CAPTCHA / reCAPTCHA\newline
    F. Flash / Flash-Lite\newline
    G. Lens\newline
    H. Google Maps\newline
    I. NVIDIA\newline
    J. Maps\newline
    K. Circle to Search\newline
    L. Gemini Live\newline
    M. Token\newline
    N. Anthropic\newline
    O. Pixel 10 / Pixel 10a\newline
    P. Alphabet\newline
    Q. YouTube\newline
    R. TikTok\newline
    S. Cutting-edge\newline
    T. ChatGPT / OpenAI\newline
    U. Advertising / Advertisement\newline
    V. Find the Look / Virtual Try-On \\
    \midrule
    Numeric prediction &
    On 2026-04-29 (UTC+8), what will Internal Three-Way Crossbred Hog hog price be, in CNY per kilogram? \\
    \bottomrule
  \end{tabular}
\end{table}

\subsection{Scoring rules}

We use type-specific scoring rules for different question formats. Only resolved questions with valid ground-truth answers are scored. If a question remains unresolved, it is excluded from scoring. As a result, the reported metrics reflect performance only on questions with confirmed outcomes.

\subsubsection{Choice-question scoring}

Suppose a choice-question has $m$ options. We denote the gold vector by $\mathbf{y}\in\{0,1\}^m$ and the prediction vector by $\hat{\mathbf{y}}\in\{0,1\}^m$, where
\begin{equation}
\begin{aligned}
y_j &=
\begin{cases}
1, & \text{if option } j \text{ is correct},\\
0, & \text{if option } j \text{ is wrong},
\end{cases}
\\[6pt]
\hat{y}_j &=
\begin{cases}
1, & \text{if option } j \text{ is selected by the agent},\\
0, & \text{if option } j \text{ is not selected by the agent}.
\end{cases}
\end{aligned}
\end{equation}
The question-level score is then computed as the option-level F1 score
\begin{equation}
S_{\mathrm{F1}}(\mathbf{y}, \hat{\mathbf{y}})
=
\frac{2\,\mathbf{y}^{\top}\hat{\mathbf{y}}}
{\|\mathbf{y}\|_1 + \|\hat{\mathbf{y}}\|_1}.
\end{equation}
For binary choice questions, the prediction must contain exactly one selected option. Any prediction that selects more than one option is treated as invalid and assigned an F1 score of 0.

\subsubsection{Numeric-question scoring}

For numeric prediction questions, we evaluate the agent relative to the recent variability of the prediction target. Let $\hat{v}$ denote the value predicted by the agent. Let
\begin{equation}
\mathcal{V} = \{v_1, v_2, \dots, v_8\}
\end{equation}
denote eight consecutive historical values associated with the prediction target, where $v_8=v$ is the resolved true value on the target date. We define the score as
\begin{equation}
S_{\mathrm{num}}
=
\max\!\left(
0,\;
1 -
\left(
\frac{\hat{v} - v}
{3\sigma(\mathcal{V}) + \varepsilon}
\right)^2
\right),
\end{equation}
where $\sigma(\mathcal{V})$ denotes the sample standard deviation of $\mathcal{V}$, and $\varepsilon$ is a small number introduced for numerical stability. This score is bounded in $\left[0,1\right]$ and drops as the prediction value moves farther from the true value, but it drops more slowly for targets that fluctuate more.

\subsubsection{Overall scoring}

\begin{table}
  \caption{Average performance of several frontier agents on the FutureWorld daily benchmark over four consecutive days. The best results are \textbf{bold-typed} and the second best ones are \underline{underlined}.}
  \label{evaluation}
  \centering
  \begin{tabular}{lccccc}
    \toprule
    \textbf{Agents} & $S_{\mathrm{bin}}$ & $S_{\mathrm{smc}}$ & $S_{\mathrm{dmc}}$ & $S_{\mathrm{num}}$ & $S_\mathrm{overall}$ \\
    \midrule
    \texttt{x-ai/grok-4.20} & 70.00 & 36.70 & 6.33 & 15.24 & 32.07 \\
    \texttt{z-ai/glm-5.1} & \underline{71.25} & 45.71 & 2.79 & 15.24 & 33.75 \\
    \texttt{openai/gpt-5.4} & \textbf{81.25} & 40.31 & \textbf{20.78} & 5.17 & 36.88 \\
    \texttt{anthropic/claude-opus-4.6} & 63.75 & \underline{47.14} & \underline{20.28} & \underline{16.84} & 37.00 \\
    \texttt{google/gemini-3.1-pro-preview} & \textbf{81.25} & 42.77 & 7.95 & \textbf{18.84} & \underline{37.70} \\
    \texttt{qwen/qwen3-max-thinking} & \textbf{81.25} & \textbf{47.38} & 15.73 & 11.70 & \textbf{39.01} \\
    \bottomrule
  \end{tabular}
\end{table}

\begin{figure}[t]
  \centering
  \includegraphics[width=1\columnwidth]{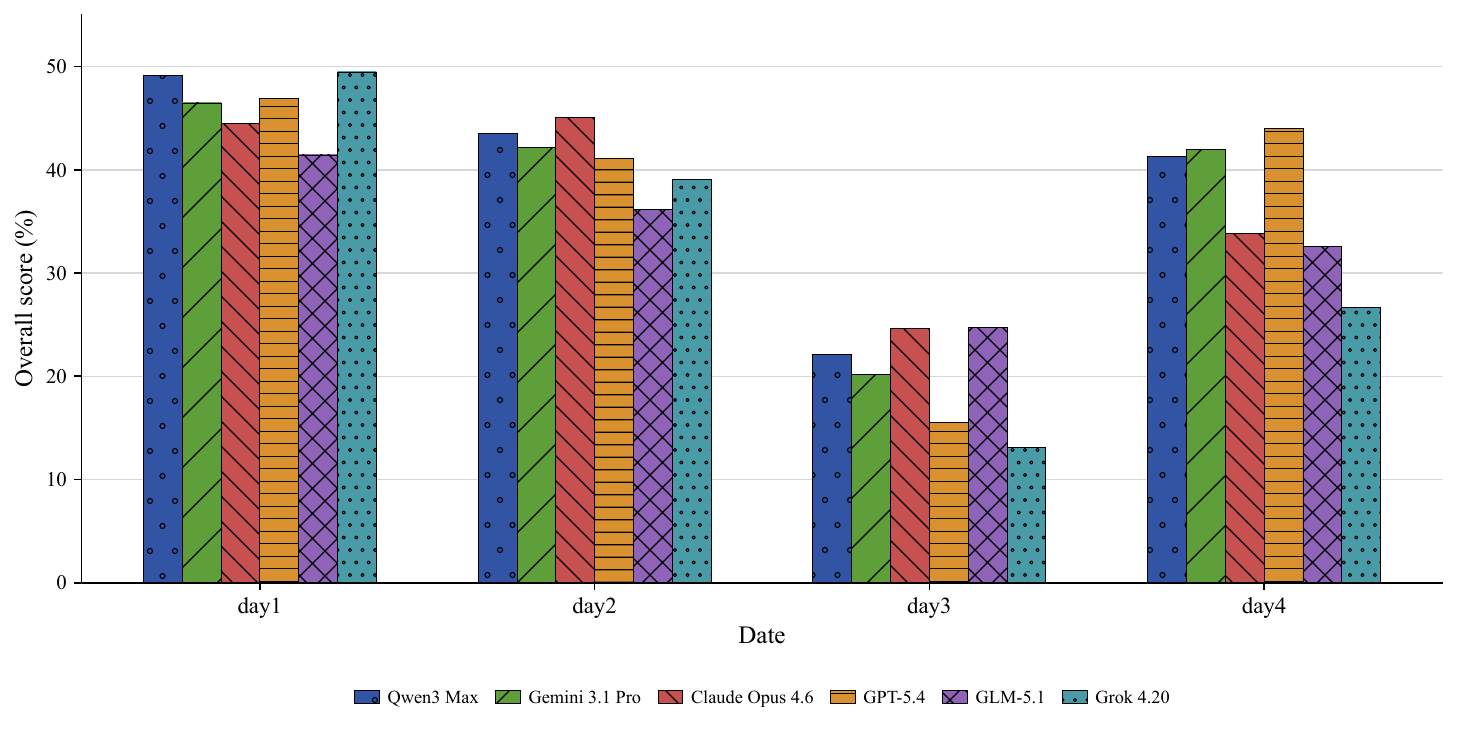}
  \caption{Overall scores of agents on the FutureWorld daily benchmark over 4 consecutive days.}
  \label{leaderboard}
\end{figure}

After computing question-level scores, we average them separately within each question type. Let $S_{\mathrm{bin}}$, $S_{\mathrm{smc}}$, $S_{\mathrm{dmc}}$, and $S_{\mathrm{num}}$ denote the mean scores for binary choice, simple multiple choice, difficult multiple choice, and numeric prediction questions, respectively. The final overall score is computed by assigning equal weight to the four question types:
\begin{equation}
S_{\mathrm{overall}}
=
\frac{1}{4}
\left(
S_{\mathrm{bin}}
+
S_{\mathrm{smc}}
+
S_{\mathrm{dmc}}
+
S_{\mathrm{num}}
\right).
\end{equation}

\subsection{Evaluation of frontier agents}

We evaluate several frontier agents\footnote{%
\begin{tabular}[t]{@{}l@{}}
\url{https://docs.z.ai/guides/llm/glm-5.1}, \url{https://grok.com/}\\
\url{https://ai.google.dev/gemini-api/docs/models/gemini-3.1-pro-preview}\\
\url{https://www.anthropic.com/news/claude-opus-4-6}
\end{tabular}%
} \cite{qwen3-max-thinking} with live web search capabilities on four consecutive days, and report the four-day average results in Table~\ref{evaluation}. On average, \texttt{qwen/qwen3-max-thinking} achieves the best overall performance. Figure~\ref{leaderboard} further shows the daily overall score achieved by each agent over the four consecutive days.

\subsection{Evaluation of open-source agent frameworks}

\begin{table}[t]
  \caption{Daily scores of agent frameworks on the FutureWorld daily benchmark, where all frameworks use GPT-5.4 as the base model. The best score is \textbf{bold-typed}.}
  \label{framework-daily-results}
  \centering
  \setlength{\tabcolsep}{4pt}
  \begin{tabular}{llccccc}
    \toprule
    \textbf{Date} & \textbf{Method} & $S_{\mathrm{bin}}$ & $S_{\mathrm{smc}}$ & $S_{\mathrm{dmc}}$ & $S_{\mathrm{num}}$ & $S_{\mathrm{overall}}$ \\
    \midrule
    \multirow{2}{*}{2026-04-25} & \texttt{smolagents} & \textbf{75.00} & \textbf{71.43} & 39.17 & 62.18 & 61.95 \\
          & \texttt{Flash-Searcher} & \textbf{75.00} & 57.14 & \textbf{60.61} & \textbf{62.35} & \textbf{63.78} \\
    \midrule
    \multirow{3}{*}{2026-04-26} & \texttt{smolagents} & \textbf{100.00} & \textbf{44.44} & 29.80 & \textbf{54.17} & 57.10 \\
          & \texttt{Flash-Searcher} & \textbf{100.00} & \textbf{44.44} & \textbf{53.94} & 47.52 & \textbf{61.48} \\
          & \texttt{OpenClaw} & 50.00 & \textbf{44.44} & 49.35 & 12.79 & 39.15 \\
    \midrule
    \multirow{4}{*}{2026-04-27} & \texttt{smolagents} & \textbf{100.00} & 57.14 & 49.26 & 43.30 & 62.43 \\
          & \texttt{Flash-Searcher} & 80.00 & 85.71 & 35.86 & \textbf{79.59} & 70.29 \\
          & \texttt{OpenClaw} & \textbf{100.00} & \textbf{100.00} & \textbf{71.11} & 52.68 & \textbf{80.95} \\
          & \texttt{Hermes} & \textbf{100.00} & 71.43 & 38.03 & 45.63 & 63.77 \\
    \midrule
    \multirow{4}{*}{2026-04-28} & \texttt{smolagents} & 80.00 & \textbf{36.67} & 41.33 & 39.83 & 49.46 \\
          & \texttt{Flash-Searcher} & 60.00 & \textbf{36.67} & 44.80 & 66.33 & 51.95 \\
          & \texttt{OpenClaw} & \textbf{100.00} & 25.00 & 28.71 & 19.78 & 43.37 \\
          & \texttt{Hermes} & \textbf{100.00} & \textbf{36.67} & \textbf{58.93} & \textbf{67.02} & \textbf{65.65} \\
    \midrule
    \multirow{4}{*}{2026-04-30} & \texttt{smolagents} & \textbf{100.00} & \textbf{60.00} & 48.21 & 53.22 & 65.36 \\
          & \texttt{Flash-Searcher} & \textbf{100.00} & 40.00 & 45.22 & -- & 61.74 \\
          & \texttt{OpenClaw} & \textbf{100.00} & 0.00 & 41.05 & 17.74 & 39.70 \\
          & \texttt{Hermes} & \textbf{100.00} & 40.00 & \textbf{50.36} & \textbf{77.45} & \textbf{66.95} \\
    \midrule
    \multirow{4}{*}{2026-05-02} & \texttt{Flash-Searcher} & \textbf{50.00} & \textbf{55.56} & 63.90 & 52.34 & \textbf{55.45} \\
          & \texttt{OpenClaw} & \textbf{50.00} & 44.44 & 43.25 & 27.23 & 41.23 \\
          & \texttt{Hermes} & 25.00 & 44.44 & \textbf{67.55} & \textbf{65.15} & 50.54 \\
          & \texttt{GPT-5.4} (web search) & \textbf{50.00} & 44.44 & 18.51 & 41.99 & 38.74 \\
    \midrule
    \multirow{5}{*}{2026-05-03} & \texttt{smolagents} & \textbf{100.00} & \textbf{44.44} & \textbf{52.80} & 48.62 & 61.47 \\
          & \texttt{Flash-Searcher} & \textbf{100.00} & \textbf{44.44} & 43.08 & 54.00 & 60.38 \\
          & \texttt{OpenClaw} & \textbf{100.00} & \textbf{44.44} & 36.92 & 24.67 & 51.51 \\
          & \texttt{Hermes} & \textbf{100.00} & \textbf{44.44} & 51.83 & \textbf{72.17} & \textbf{67.11} \\
          & \texttt{GPT-5.4} (web search) & 75.00 & \textbf{44.44} & 19.64 & 50.64 & 47.43 \\
    \midrule
    \multirow{4}{*}{2026-05-04} & \texttt{smolagents} & \textbf{100.00} & \textbf{62.50} & 50.95 & 43.16 & 64.15 \\
          & \texttt{Flash-Searcher} & \textbf{100.00} & \textbf{62.50} & \textbf{69.05} & \textbf{59.00} & \textbf{72.64} \\
          & \texttt{Hermes} & 60.00 & 50.00 & 21.94 & 23.30 & 38.81 \\
          & \texttt{GPT-5.4} (web search) & 40.00 & 50.00 & 4.76 & 19.82 & 28.64 \\
    \midrule
    \multirow{5}{*}{2026-05-05} & \texttt{smolagents} & 20.00 & 14.29 & 36.11 & 45.31 & 28.93 \\
          & \texttt{Flash-Searcher} & \textbf{60.00} & 7.14 & \textbf{48.61} & \textbf{60.60} & \textbf{44.09} \\
          & \texttt{OpenClaw} & \textbf{60.00} & \textbf{21.43} & 43.19 & 19.36 & 36.00 \\
          & \texttt{Hermes} & \textbf{60.00} & \textbf{21.43} & 42.22 & 31.97 & 38.90 \\
          & \texttt{GPT-5.4} (web search) & 40.00 & 0.00 & 12.50 & 34.37 & 21.72 \\
    \bottomrule
  \end{tabular}
\end{table}

We additionally evaluate agents built with several open-source agent frameworks on the FutureWorld daily benchmark. The evaluated frameworks cover different agent-system design choices. \texttt{smolagents} is a lightweight Hugging Face library\footnote{\url{https://huggingface.co}} that exposes a compact interface for building tool-using agents and connecting language models with external tools and code execution \cite{hf2024smolagents}. \texttt{Flash-Searcher} organizes web search and evidence reading as a DAG-based parallel execution process, aiming to reduce serial search latency and improve evidence coverage \cite{qin2025flash}. \texttt{OpenClaw} is a local personal-assistant framework with a unified execution gateway, multi-agent routing, and skill-extension mechanisms \cite{openclaw2026}. \texttt{Hermes} emphasizes long-term memory and self-improvement, including cross-session retrieval, summary, user modeling, skill generation, and parallel sub-agent execution \cite{hermesagent2026}. Table~\ref{framework-daily-results} summarizes the performance of different agent frameworks, where all frameworks use GPT-5.4 as the base model.

\section{Prompt templates for FutureWorld daily benchmark}
\label{templates}

The following templates are used to construct prompts for the four question formats in the FutureWorld daily benchmark. In each template, \texttt{<QUESTION>} is replaced by the concrete prediction question.

\subsection{Binary choice}

\begin{lstlisting}[style=prompttemplate]
You are an agent that can predict future events. The event to be predicted:
"""
<QUESTION>
"""

Your goal is to identify the single correct option based on your analysis.

NOTE: If you notice a potential time conflict in the question, do not worry—this may be due to different time zones, and it can still refer to the same moment in time.

IMPORTANT: Your final answer MUST end with this exact format:
\boxed{A} or \boxed{B}

Do not use any other format. Do not refuse to make a prediction. Do not say "I cannot predict the future." You must make a clear prediction based on the best data currently available, using the box format specified above.
\end{lstlisting}

\subsection{Simple multiple choice}

\begin{lstlisting}[style=prompttemplate]
You are an agent that can predict future events. The event to be predicted:
"""
<QUESTION>
"""

Your goal is to identify all the correct option(s) based on your analysis. There may be exactly one correct option or multiple correct options; you must determine the correct set.

Please list all correct option(s) you have identified, separated by commas. If you believe there is only one correct option, output that option alone (without any commas).

Output only the option label(s). Do not output the full option text. If you output the option text in addition to the option label(s), your answer will be marked incorrect even if you chose the correct option(s).

NOTE: If you notice a potential time conflict in the question, do not worry—this may be due to different time zones, and it can still refer to the same moment in time.

IMPORTANT: Your final answer MUST end with this exact format:
\boxed{label 1} or \boxed{label 2, label 3, ...}

For example: \boxed{A} for a single correct option, or \boxed{B, C} for multiple correct options.

Do not use any other format. Do not refuse to make a prediction. Do not say "I cannot predict the future." You must make a clear prediction based on the best data currently available, using the box format specified above.
\end{lstlisting}

\subsection{Difficult multiple choice}

\begin{lstlisting}[style=prompttemplate]
You are an agent that can predict future events. The event to be predicted:
"""
<QUESTION>
"""

Your goal is to identify all the correct option(s) based on your analysis. There may be exactly one correct option or multiple correct options; you must determine the correct set.

Please list all correct option(s) you have identified, separated by commas. If you believe there is only one correct option, output that option alone (without any commas).

Output only the option label(s). Do not output the full option text. If you output the option text in addition to the option label(s), your answer will be marked incorrect even if you chose the correct option(s).

NOTE: If you notice a potential time conflict in the question, do not worry—this may be due to different time zones, and it can still refer to the same moment in time.

IMPORTANT: Your final answer MUST end with this exact format:
\boxed{label 1} or \boxed{label 2, label 3, ...}

For example: \boxed{A} for a single correct option, or \boxed{B, C} for multiple correct options.

Do not use any other format. Do not refuse to make a prediction. Do not say "I cannot predict the future." You must make a clear prediction based on the best data currently available, using the box format specified above.
\end{lstlisting}

\subsection{Numeric prediction}

\begin{lstlisting}[style=prompttemplate]
You are an agent that can predict future events. The event to be predicted:
"""
<QUESTION>
"""

Your goal is to make a numeric prediction.

NOTE: If you notice a potential time conflict in the question, do not worry—this may be due to different time zones, and it can still refer to the same moment in time.

IMPORTANT: Your final answer MUST end with this exact format:
\boxed{number}

Do not use any other format. Do not refuse to make a prediction. Do not say "I cannot predict the future." You must make a clear prediction based on the best data currently available, using the box format specified above.

Your output must be a plain numeric value only. Do NOT include any units, currency symbols, percent signs, commas, spaces, scientific notation (e.g., 1e6), or any other extra characters. Only digits are allowed, with an optional leading minus sign and an optional decimal point (e.g., 123, -45, 67.89).
\end{lstlisting}

\section{Broader impacts}

FutureWorld may have positive societal impacts by supporting the development of predictive agents that learn from real-world feedback and adapt to changing conditions. Such agents may help improve decision support in domains where anticipating future events is useful, such as public-interest monitoring, scientific analysis, operations planning, and risk assessment.

At the same time, stronger predictive agents may also create risks if used inappropriately. For example, they could be used to support speculative trading, political targeting, or other decisions that over-rely on uncertain predictions. Incorrect predictions may also mislead users if their uncertainty is not properly communicated. To mitigate these risks, FutureWorld focuses on probabilistic prediction rather than deterministic claims, filters out sensitive or otherwise inappropriate questions, and discusses possible advantages from prediction-market and odds-aggregation websites. We also apply safeguards for responsible release. FutureWorld only uses publicly accessible sources and filters out sensitive, harmful, or otherwise inappropriate questions.

\section{Existing assets, licenses, and terms of use}

We use several existing assets, including open-source code packages, pretrained models, APIs, and public online data sources. We cite the original creators where applicable and follow the corresponding licenses or terms of use. Table~\ref{assets} summarizes the main existing assets used in this work.

\begin{table}[t]
  \centering
  \footnotesize
  \caption{Assets used in this work.}
  \label{assets}

  \begingroup
  \renewcommand{\tabularxcolumn}[1]{m{#1}} 
  \renewcommand{\arraystretch}{1.08}

  \begin{tabularx}{\linewidth}{
    L{0.30\linewidth}
    L{0.22\linewidth}
    Y}
    \toprule
    \textbf{Asset} & \textbf{License / terms} & \textbf{Usage} \\
    \midrule
    \texttt{verl-tool} &
    MIT License &
    Used as the base tool-agent RL framework and modified into \texttt{verl-tool-future}. \\

    PyTorch FSDP &
    BSD-style / BSD-3-Clause &
    Used as the distributed training backend. \\

    vLLM &
    Apache-2.0 License &
    Used for efficient rollout generation. \\

    FlashAttention-2 &
    BSD-3-Clause License &
    Used to accelerate attention computation during training and rollout. \\

    \texttt{Qwen3-4B-Instruct-2507} &
    Apache-2.0 License &
    Used as one of the open-source base models for training predictive agents. \\

    \texttt{Qwen2.5-3B-Instruct} &
    Qwen Research License &
    Used as one of the open-source base models for training predictive agents. \\

    \texttt{DeepSeek-R1-0528-Qwen3-8B} &
    MIT License &
    Used as one of the open-source base models for training predictive agents. \\

    \texttt{BAAI/bge-small-en-v1.5} &
    MIT License &
    Used to encode question-description pairs for semantic resampling. \\

    Serper API &
    Serper API terms of service &
    Used to provide the Google-search interface during agent rollouts. \\

    \texttt{AkShare} &
    MIT License &
    Used to retrieve part of the outcome data. \\

    Public source websites &
    Source-specific access conditions &
    Used to collect candidate future events from publicly accessible sources. \\
    \bottomrule
  \end{tabularx}

  \endgroup
\end{table}



\end{document}